\documentclass[10pt,twocolumn,letterpaper]{article}

\usepackage{iccv}
\usepackage{graphicx}
\usepackage{amsmath}
\usepackage{amsthm}
\usepackage{booktabs}
\usepackage{algorithm}
\usepackage{algorithmic}
\usepackage[switch]{lineno}
\usepackage{multirow}
\usepackage[para]{footmisc}

\usepackage[pagebackref=true,breaklinks=true,letterpaper=true,colorlinks,bookmarks=false]{hyperref}

\iccvfinalcopy 

\def\iccvPaperID{1} 

\begin{document}

\title{Layer Pruning with Consensus: A Triple-Win Solution}

\author{
	Leandro Giusti Mugnaini\thanks{These authors contributed equally to this work.} \and
	Carolina Tavares Duarte\footnotemark[1] \and
	Anna H. R. Costa \and
	Artur Jordao
}

\author{
	Leandro Giusti Mugnaini\thanks{These authors contributed equally to this work.} \quad
	Carolina Tavares Duarte\footnotemark[1] \quad
	Anna H. R. Costa,
	Artur Jordao \\
	Escola Politécnica, Universidade de São Paulo, Brasil \\
}

\maketitle

\begin{abstract}
Layer pruning offers a promising alternative to standard structured pruning, effectively reducing computational costs, latency, and memory footprint. While notable layer-pruning approaches aim to detect unimportant layers for removal, they often rely on single criteria that may not fully capture the complex, underlying properties of layers. We propose a novel approach that combines multiple similarity metrics into a single expressive measure of low-importance layers, called the Consensus criterion. Our technique delivers a triple-win solution: low accuracy drop, high-performance improvement, and increased robustness to adversarial attacks. With up to 78.80\% FLOPs reduction and performance on par with state-of-the-art methods across different benchmarks, our approach reduces energy consumption and carbon emissions by up to 66.99\% and 68.75\%, respectively. Additionally, it avoids shortcut learning and improves robustness by up to 4 percentage points under various adversarial attacks. Overall, the Consensus criterion demonstrates its effectiveness in creating robust, efficient, and environmentally friendly pruned models.
keywords{Layer Pruning  \and Similarity Metric \and Robustness.}
\end{abstract}
\section{Introduction}\label{sec:introduction}

Deep learning is advancing machine learning toward human-level performance in many cognitive tasks such as computer vision and natural language processing~\cite{Touvron:2023}.
In this direction, over-parameterized models have gained popularity for their ability to represent high-complexity patterns in data, making it easier to solve non-convex problems.
On the other hand, such models suffer from high computational costs and memory consumption, hindering their applicability in low-resource and infrastructure-less scenarios. Additionally, high-capacity models are prone to incorrect predictions under adversarial attacks — small perturbations in the input that force a model to make mistakes in its predictions — making them unreliable for safety- and security-critical tasks~\cite{Hendrycks:2019,Yang:2024,DBLP:journals/pami/PeckGS24}. These issues pose the following dilemma: \emph{how to obtain high-predictive ability, low-cost and robust models?}
\begin{figure*}[!tb]
	\centering
	\includegraphics[width=0.45\linewidth]{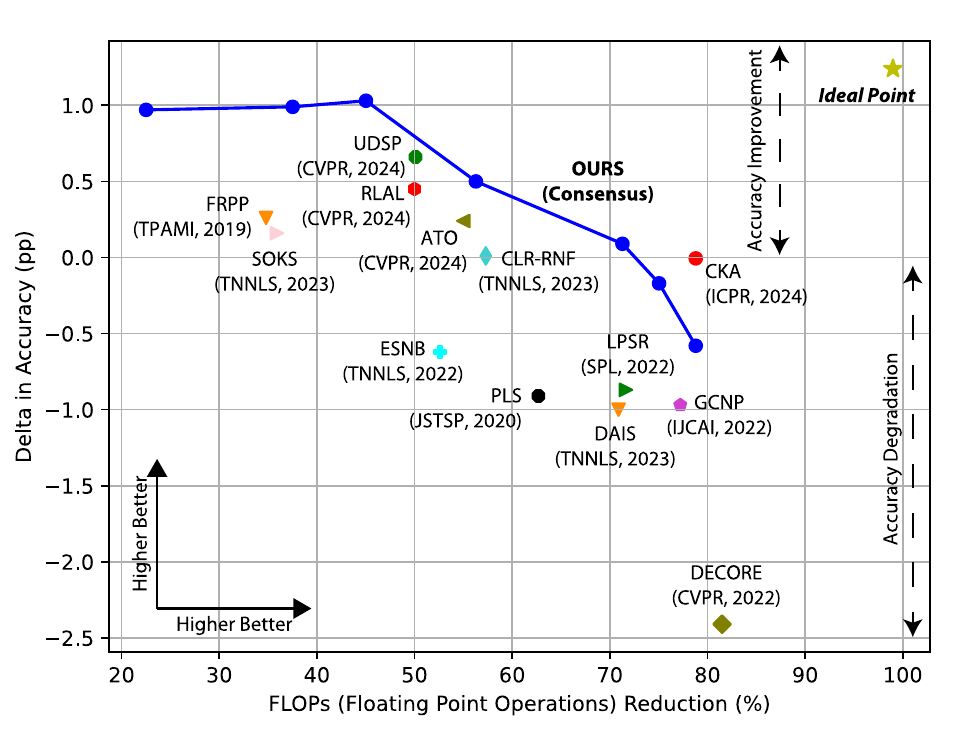}
	\includegraphics[width=0.5\linewidth]{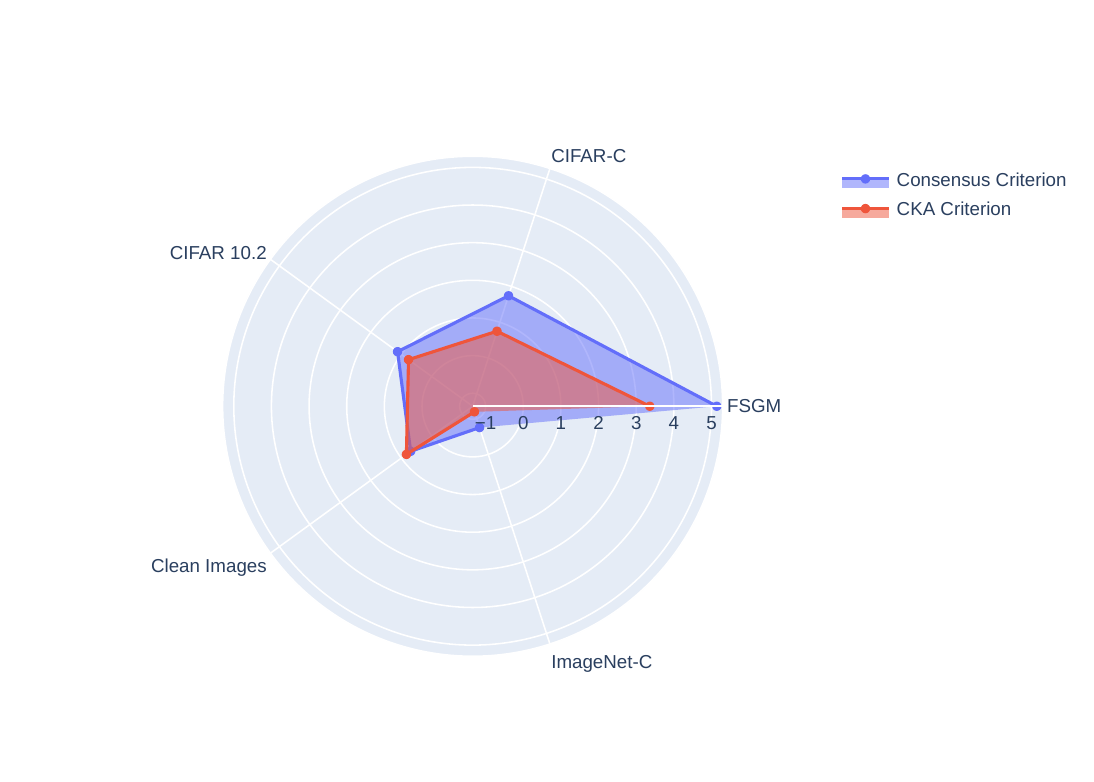}
	\caption{Left: Comparison with state-of-the-art on the popular ResNet56 + CIFAR-10 setting. Our Consensus technique achieves one of the best trade-offs between accuracy and computational reduction (estimated by FLOPs). Compared to state-of-the-art pruning techniques, our method reduces FLOPs by up to 71\% without compromising model accuracy. 
		Right: Comparison between our method and a recent state-of-the-art layer-pruning criterion: CKA~\cite{Pons:2024}. In this evaluation, we report the mean accuracy delta between multiple pruning iterations and the original, unpruned, model. We evaluate the models on well-known adversarial and Out-Of-Distribution (OOD) benchmarks, as well as on clean images to assess generalization. Our Consensus criterion exhibits increased robustness to adversarial attacks and OOD samples, while maintaining generalization on clean images.}
	\label{fig:teaser}
\end{figure*}

Existing studies confirm that pruning strategies emerge as promising solutions to address the aforementioned dilemma~\cite{Bartoldson:2020,Jordao:2021,Phan:2023,Bair:2024}. For example, state-of-the-art pruning techniques remove more than $75\%$ of floating point operations (FLOPs) and parameters without compromising model accuracy~\cite{Pons:2024, He:2023}. 
This family of techniques also exhibit positive results in improving adversarial robustness, even when traning only on clean images or adversarial samples~\cite{Bartoldson:2020,Jordao:2021,Phan:2023,Bair:2024}.
Such benefits have attracted intense research on pruning techniques and confirm that it occupies an important place in the era of foundation models~\cite{Ma:2023,Sun&Wanda:2024}.

The spectrum of pruning techniques includes unstructured and structured approaches~\cite{He:2023,Cheng:2024:PAMI}. While the first focuses on removing individual weights, the latter removes entire neurons (i.e., filters) from the model.
Structured pruning is widely recognized as being more hardware-friendly~\cite{Bair:2024,Shen:2022}, as its unstructured counterpart often requires technologies capable of handling sparse matrix computations to achieve practical gains.

Recent efforts in structured pruning have focused on eliminating large structures, such as layers or entire building blocks~\cite{Pons:2024, Jordao:2020,  Chen:2019, Zhou:2022, Zhang_Liu:2022, Xia:2024, Kim:2024}. It turns out that removing small structures, such as neurons or filters, is less effective in reducing the latency (inference time) of a model~\cite{Pons:2024, Zhou:2022, Xia:2024, Kim:2024}. In particular, according to early studies~\cite{Dehghani:2022,Vasu:2023}, common metrics such as FLOPs and the number of parameters exhibit a weak correlation with latency. Conversely, layer pruning maintains all the benefits of small structure pruning and also offers additional advantages~\cite{Pons:2024,Zhou:2022}.

In the context of layer pruning, modern techniques frequently extend simple filter criteria (e.g., averaging $\ell_1$-norm) to assign layer importance and, then, remove the least important ones~\cite{Jordao:2020,Zhang:2022}. Pons et al.~\cite{Pons:2024} confirmed that these strategies do not adequately capture the underlying properties of layers and, hence, hurt model accuracy during pruning (mainly at high compression rates). To address this problem, the authors proposed to apply the Centered Kernel Alignment (CKA) metric as a pruning criterion. 
Despite positive results on generalization (i.e., evaluation on clean images — i.i.d. samples), the pruned models obtained by the CKA criterion exhibit suboptimal robustness against adversarial samples.

%
Our method, called Consensus, is built upon the ideas by Pons et al.~\cite{Pons:2024}, but here, instead of relying solely on CKA, we integrate multiple similarity metrics.
Our main motivation relies on the fact that CKA does not satisfy the triangle inequality leading to pitfalls in capturing representational differences~\cite{Duong:2023,Williams:2021}.
Additionally, using a single criterion can result in models that appear to perform well on i.i.d samples but fail in more challenging situations, such as adversarial attacks and Out-of-Distribution (OOD) data. This is because relying on a single criterion can lead to similar and potentially biased patterns in ranking the importance of structures, making it difficult to differentiate between truly important and less important structures. Consequently, this undermines the confidence in these rankings, as the single criterion may not provide a comprehensive and accurate assessment of layer importance~\cite{Huang&Shao:2021}.

Additionally, following the shortcut learning phenomenon~\cite{Geirhos:2020, Hermann:2023}, we believe that pruned models of a single criterion may exploit shortcut opportunities from clean images, learning discriminative representations that struggle to generalize on more challenging scenarios (i.e., adversarial attacks and OOD samples). Shortcut learning occurs when models rely on unintended features or shortcuts that work well on standard benchmarks but fail under different conditions~\cite{Geirhos:2020, Hermann:2023}. This discrepancy between intended and actual learning strategies often leads to suboptimal predictive ability in OOD scenarios. In this direction, our results suggest that our criterion yields pruned models that avoid shortcut opportunities better than the most promising criterion for layer pruning, as our models achieve superior trade-offs in generalization and robustness (see Figure~\ref{fig:teaser}, right).

To compose our Consensus criterion, we use similarity metric spaces for stochastic neural networks proposed by Duong et al.~\cite{Duong:2023}. Powered by these metrics, the primary contribution of this paper is demonstrating that combining multiple similarity metrics as a single pruning criterion results in more robust and trustworthy pruned models. 
Thus, the combination of multiple similarity metrics as a pruning criterion offers several benefits. The following triad stands out:

\noindent
\textbf{Low Accuracy Drop}: By integrating multiple similarity metrics, our pruning method maintains high accuracy even at high compression rates. The comprehensive evaluation of layer importance reduces the risk of removing critical layers that are essential for maintaining performance on clean images.

\noindent
\textbf{High-Performance Improvement}: Our approach significantly reduces computational costs by effectively pruning less important layers. This leads to substantial improvements in inference time and memory consumption, making the models more suitable for deployment in low-resource environments.

\noindent
\textbf{Increased Robustness to Adversarial Attacks}: The use of multiple similarity metrics provides a more robust assessment of layer importance, enhancing the model's resilience against adversarial attacks. By avoiding dependence on a single metric, our method ensures that pruned models can withstand various adversarial perturbations, improving their reliability in safety-critical applications.

By combining multiple similarity metrics, our approach provides a more reliable and robust measure of layer importance, addressing the limitations of single-metric methods. Although our method is versatile to other forms of pruning, we focus on removing layers as previous works have confirmed its benefits across all standard computational metrics and beyond~\cite{Pons:2024, Zhou:2022,Xia:2024,Kim:2024}, as illustrates Figure~\ref{fig:teaser}.

Extensive experiments demonstrate the effectiveness of our approach. Specifically, on CIFAR-10 and ImageNet, our method achieves state-of-the-art performance in terms of accuracy drop and FLOPs reduction. For example, at a FLOP reduction of up to $78.8\%$, our method exhibits low accuracy drop, outperforming existing techniques (see Figure~\ref{fig:teaser}). Additionally, our approach significantly enhances adversarial robustness, as evidenced by its superior performance on multiple adversarial benchmarks, such as CIFAR-10.2, CIFAR-C, FGSM, and ImageNet-C~\cite{Hendrycks:2019, Lu:2020}.

In summary, our layer-pruning technique~\footnotetext[1]{\parbox{\columnwidth}{Code is available at: \url{https://github.com/CarolinaTavaresDuarte/Consensus-Layer-Pruning/}}}
surges as a triple-win solution: low accuracy drop, high-performance improvement and increased robustness to adversarial attacks.

\section{Related Work}\label{sec:related}
Researchers have intensely focused on pruning methods to reduce model complexity and computational resources. These techniques are crucial for making high-performance models more accessible in low-resource environments~\cite{He:2023,Cheng:2024:PAMI}.

\noindent
\textbf{Unstructured Pruning Methods.}
Out-of-the-shelf pruning methods often rely on criteria such as the magnitude of weights to identify and zero out unimportant weights (unstructured pruning)~\cite{Kwon:2022,HuanWang:2022}.
While effective in reducing model size, these methods face limitations, such as low variance in importance scores and difficulties in comparing norms across different regions of the architecture~\cite{Huang&Shao:2021,Jordao:2023,Zhang:2022}.

Additionally, in the context of LLMs, Sun et al.~\cite{Sun&Wanda:2024} observed that \(\ell_p\)-norm criteria fail to capture unimportant structures when the input varies significantly in scale. To mitigate this, the authors proposed to project a few samples into the norm to measure prunable weights. Their method belongs to unstructured pruning; therefore, it requires specialized hardware for sparse computing. For this purpose, the authors extended their algorithm to employ the N:M group pruning and take advantage of recent NVIDIA’s sparse tensor (i.e., A100)~\cite{Zhou:2021}.

Efforts have been dedicated to the study of more elaborated criteria for unstructured pruning. For example, Frantar and Alistarh~\cite{Frantar:2023} use a sparse regression solver to remove weights based on row-wise Hessian reconstruction.

In contrast to the above efforts, our layer-pruning method offers computational benefits without requiring specific hardware or software. In addition, due to the similarity metrics our criterion employs, it can compare LLM representations; therefore, our Consensus criterion is adaptable to this family of models. However, this exploration is beyond our current scope.

\noindent
\textbf{Structured Pruning Methods.}
Apart from the unstructured pruning, recent advancements have shifted focus toward eliminating large structures like layers~\cite{Jordao:2020, Chen:2019, Zhou:2022, Zhang_Liu:2022, Dror:2022, Fu&Yang:2022}. Pruning layers not only retains the benefits of structural pruning but also reduces latency~\cite{Pons:2024, Zhou:2022}.
For example, Dror et al.~\cite{Dror:2022} and Fu et al.~\cite{Fu&Yang:2022} use structural reparameterization to merge layers, reducing model depth and addressing the issue of low variance in importance scores.
The method by Liu et al.~\cite{Liu:2024} combines a progressive training strategy with block pruning, balancing the trade-offs between depth reduction and performance. However, their technique strongly relies on neural architecture search, making it less directly comparable to existing layer pruning methods, including our own.

More similar to our work, Pons et al.~\cite{Pons:2024} observed that extending weights or pruning criteria for scoring layers is inadequate since they do not capture the underlying properties of large structures composing the network. For this purpose, the authors proposed to employ the Centered Kernel Alignment (CKA) metric to identify and remove unimportant layers.
While Pons et al.~\cite{Pons:2024} demonstrated the effectiveness of CKA in maintaining model generalization, CKA criterion exhibits suboptimal robustness against adversarial samples.

Unlike the aforementioned approaches that rely on a single metric, our method combines several similarity metrics to form a comprehensive criterion for identifying low-importance layers. This approach addresses the limitations of single-metric methods and enhances the overall performance of pruned models. Therefore, our novel criterion effectively identifies unimportant layers, surpassing existing layer-pruning methods and other state-of-the-art pruning techniques.

\noindent
\textbf{Pruning as a Form of Adversarial Defense.}
Besides computational challenges, another major concern with deep models is adversarial attacks. These attacks pose a threat to the reliability of deep learning models, especially for safety- and security-critical tasks~\cite{Yang:2024,Cheng:2024:PAMI}.

Techniques such as adversarial training and data augmentation methods aim to enhance robustness but often come with substantial computational overhead, particularly in the training phase~\cite{Hendrycks:2022}. 
Surprisingly, early works confirmed that pruned models exhibit adversarial robustness under certain conditions~\cite{Bartoldson:2020,Jordao:2021,Phan:2023,Bair:2024}.

Particularly, structured pruning enhances robustness against adversarial attacks by simplifying model complexity. Mitra et al.~\cite{Mitra:2024} explored robustness to natural corruption and uncertainty calibration in post-hoc pruned models, finding that pruning significantly enhances uncertainty calibration and can maintain or improve robustness to natural corruption compared to unpruned models. Furthermore, Li et al.~\cite{Li:2023} demonstrated that pruning enhances certified robustness by reducing neuron instability and tightening verification bounds. These evidence underscore that pruning and adversarial defense mechanisms are orthogonal, allowing for their combination to yield even more robust models, improving the safety and reliability of neural networks in practical tasks.
The Consensus technique fosters this field by combining multiple similarity metrics to identify and remove low-importance layers. Our approach not only enhances computational efficiency but also improves robustness against adversarial attacks, addressing the limitations of existing pruning methods and contributing significantly to the development of more efficient and reliable deep learning models.
\section{Methodology}
\noindent
\textbf{Problem Statement.}
Following previous works~\cite{Pons:2024,Jordao:2020,Zhou:2022,Zhang_Liu:2022}, our goal is to identify and eliminate non-essential layers while mantaining the model's predictive ability, even at high compression rates. This approach is grounded in two key principles: (I) the residual connections within residual-based architectures enable information to flow  through multiple paths within the network~\cite{Veit:2016, Huang:2016, Dong:2021}, suggesting that layers may not always strongly depend on each other, thus reinforcing the idea of redundancy between structures; (II) a limited subset of layers is critical to the overall performance of the network~\cite{Zhang:2022,Masarczy:2023}. Based on these principles, given a network $\mathcal{F}$ composed of a layer set $L$, our goal is to remove layers to derive a shallower network $\mathcal{F'}$ with a reduced set $L'$, where $|L'|$ \(\ll\) $|L|$.
Compared to the unpruned network $\mathcal{F}$, we expect the pruned network $\mathcal{F'}$ to exhibit three key characteristics: low accuracy drop, high-performance improvement and an increased robustness to adversarial attacks.

\noindent
\textbf{Definitions.} 
Let $X$ denote training samples, such as images, and $Y$ their corresponding class labels. 
We denote $\mathcal{F}$ as a dense, unpruned, network trained using supervised learning on $X$ and $Y$. Assume $M(\cdot, X)$ is a function that extracts the feature representations from a given model using the samples $X$. Following Evci et al.~\cite{Evci:2022,Kirichenko:2023}, $M$ extracts feature maps from the layer directly before the classification layer. These feature maps encapsulate both the spurious and relevant features~\cite{Kirichenko:2023}, corresponding to a high fidelity representation of the entire network. Let $l \in L$ denote a potential layer for pruning and $S$ a set of similarity metrics. We denote the pruned network resulting from the removal of layer $l$ from $\mathcal{F}$, using the similarity metric $s$ ($s \in S$), as $\mathcal{F}_{l}^{s}$.

\noindent
\textbf{Similarity Metrics.}
The work of Pons et al.~\cite{Pons:2024} introduces a layer pruning criterion based on the CKA metric to measure the similarity between the representations of the original neural network and a candidate layer for pruning. Their study, however, is limited to the CKA metric while our Consensus criterion adopts a more comprehensive approach by integrating a set $S$ of similarity metrics to guide the layer-pruning process. Our empirical analysis suggests that using this set avoids the limitations of relying on a single metric, such as the potential for shortcut learning opportunities.
For this purpose, we consider the metrics developed by Duong et al.~\cite{Duong:2023}. The authors leverage common distance measures, such as Procrustes and Bures distance, to create a new set of metrics capable of comparing stochastic representations of neural networks. When using these metrics, we take into account the structure and scale of noise in neural responses, an important detail that deterministic metrics often overlook. Our method also inherits key properties from these shape metrics, making it invariant to rotations and effective at measuring distances in high-dimensional spaces, thus overcoming the limitations of traditional distance metrics. 
We also explore interpolated versions of these metrics that balance the penalization of differences in mean or covariance, offering a more comprehensive view of neural representations~\cite{Huang&Shao:2021, He&Huang:2021}. 

Altogether, the above metrics allow us to create a unified and robust measure of low-importance layers, as we explain below.

\noindent
\textbf{Proposed Method.}
For each similarity metric $s \in S$ and layer $l \in L$, we obtain $\mathcal{F}_{l}^{s}$ as previously defined, and apply $M(\mathcal{F}_{l}^{s}, X)$ to extract its representation, denoted by $R_{l}^{s}$. The metric $s(\cdot, \cdot)$ takes $R$ and $R_{l}^{s}$, where $R \leftarrow M(\mathcal{F}, X)$ (i.e., the representation from the unpruned model), and outputs the score of $l$.

\begin{algorithm}[!b]
	\small
	\caption{Layer Pruning iteration using our Consensus criterion}
	\label{alg::pruning}
	\begin{algorithmic}[1]
		\item[] \textbf{Input:} Trained Neural Network $\mathcal{F}$, Candidate Layers $l \in L$ \, Training Samples $X$, Similarity Metrics $S$
		\item[] \textbf{Output:} Pruned Version of $\mathcal{F}$\\
		\STATE $R \leftarrow M(\mathcal{F}, X)$ $\triangleright$ Representation extraction of $\mathcal{F}$
		\FOR{$s$ {\bfseries in} $S$}
		\FOR{$l$ {\bfseries in} $L$}
		\STATE $\mathcal{F}_{l}^{s} \leftarrow \mathcal{F} \setminus l $ $\triangleright$ Removes layer $l$ from $\mathcal{F}$
		\STATE $R_{l}^{s} \leftarrow M(\mathcal{F}_{l}^{s}, X)$  $\triangleright$ Representation extraction of $\mathcal{F}_{l}^{s}$
		\STATE $D \leftarrow D \cup s(R, R_{l}^{s})$ $\triangleright$ Similarity value of layer $l$ w.r.t the similarity metric $s$
		\ENDFOR
		\STATE $D \leftarrow ranked(D)$  $\triangleright$ Adds the ranking information for each layer using the similarity value
		\STATE $D \leftarrow sorted(D)$  $\triangleright$ Sorts the layers using the layer index
		\STATE $T_s \leftarrow D$
		\ENDFOR
		\FOR{$l$ {\bfseries in} $L$}
		\FOR{$s$ {\bfseries in} $T$}
		\STATE $V_{l} = V_{l}+V_{l_{s}[ranking]} $  $\triangleright$ Sums the ranking value from each metric $s$ for the layer $l$
		\ENDFOR	
		\ENDFOR	
		\STATE $V \leftarrow sorted(V)$  $\triangleright$ Sorts the layers using the ranked similarity values 
		\STATE $n \leftarrow argmin(V)$  $\triangleright$ Gets the first layer in the sorted ranking (most similar layer)
		\STATE \label{line10} $\mathcal{F} \leftarrow \mathcal{F}_{l_n}$ $\triangleright$ $\mathcal{F}$ becomes its pruned version without layer $l$\
		\STATE Update $\mathcal{F}$ via standard supervised paradigm on $X$
	\end{algorithmic}
\end{algorithm}

Since the similarity metrics have different score magnitudes, for a fair comparison between the importance of metrics, we sort the layers using the score and assign each layer a numerical ranking based on its position. In this scenario, the first layer is the most similar (receiving a ranking of 1) and the last layer is the least similar (receiving a ranking of $|L|$).
Then, for each $l \in L$, we sum the respective rankings from each metric and use the result as the final Consensus score. Finally, we remove the  layer with the lowest score from $\mathcal{F}$. In other words, we remove the layer that yields a representation with the highest similarity compared with the unpruned network representation.
The intuition behind this process is that by removing the most similar layer, we preserve the internal representation of the model and, thus, retain the underlying information.

Following the layer pruning literature~\cite{Pons:2024, Jordao:2020}, we conduct an iterative process to obtain pruned models with varying compromises between accuracy drop and FLOP reduction.
%
After each pruning iteration, we conduct the common approach of fine-tuning the model in order to preserve its predictive ability~\cite{Pons:2024,Williams:2024}.
Algorithm~\ref{alg::pruning} summarizes the process of a single pruning iteration using the Consensus criterion.

We highlight that the Consensus criterion also works for filter pruning. However, previous works demonstrated the benefits of layer pruning over filter pruning~\cite{Pons:2024, Jordao:2020,Zhou:2022}. It turns out that layer pruning reduces network depth, directly addressing model latency and significantly speeding up the training/fine-tuning stages. Additionally, it also provides benefits of filter pruning, such as reductions in FLOPs, memory footprint and carbon emission. Therefore, we focus on this form of pruning and leave exploring our method in filter pruning for future research.
%
\section{Experiments}\label{sec:experiments}

\noindent
\textbf{Experimental Setup.}
We conduct experiments on CIFAR-10 and ImageNet using different versions of the ResNet architecture in order to evaluate the pruning effectiveness of the Consensus criterion. Throughout both training and fine-tuning phases, we follow Pons et al.~\cite{Pons:2024} and apply random crop and horizontal flip as data augmentation. This approach ensures that improvements stem from pruning itself, rather than from additional techniques such as adversarial training or powerful data augmentations. 

Following previous works~\cite{Pons:2024, Jordao:2021}, we employ model-specific and agnostic attacks to evaluate the adversarial robustness of the Consensus criterion. For the first, we employ the Fast Gradient Sign Method (FGSM). For the latter, we use the CIFAR-10.2, CIFAR-C and ImageNet-C datasets~\cite{Hendrycks:2019, Lu:2020}. We consider the highest level of severity to the semantic-preserving attacks ($severity=4$ to CIFAR-C and $severity=5$ to ImageNet-C) and $\epsilon=16/255$ to FGSM.

To assess the predictive ability of the unpruned models against their pruned versions, we adhere to standard practices by reporting the difference in accuracy~\cite{He:2023,Williams:2024}. For the semantic-preserving attacks CIFAR-C and ImageNet-C, we report the average across all possible attacks as suggested by previous work~\cite{Hendrycks:2019}. Regardless of the dataset, negative values mean a decrease in accuracy, while positive values denote an improvement, both measured in percentage points (pp).

\noindent
\textbf{Comparison with the State of the Art.}
We start our experiments by comparing the proposed method against top-performing pruning techniques. For this purpose, we consider representative filter and layer pruning methods based on the survey by He et al.~\cite{He:2023}.
For a fair comparison, we report the results of each method according to the original paper.

Tables \ref{tab:main_resultsResNet56_left} and \ref{tab:main_resultsResNet56_rigth} summarizes the results. 
%
%
On CIFAR-10 with ResNet56, our method achieves state-of-the-art performance in both terms of accuracy drop and FLOPs reduction. For example, at a FLOP reduction of up to $60\%$, our method obtains one of the best tradeoffs between delta in accuracy and FLOP reduction. Notably, it achieves a FLOPs reduction twice as large as other criteria while simultaneously improving accuracy. At FLOP reduction above $70\%$, we observe a similar behavior jointly with CKA. Figure~\ref{fig:teaser} (left) reinforces these results, showing that our method is on par with (and often outperforms) existing state-of-the-art techniques. Finally, at the highest FLOP reduction achievable by pruning layers ($78.80\%$), our method preserves accuracy better than CKA~\cite{Pons:2024}.
\begin{table}[H]
	\caption{Comparison with state-of-the-art pruning methods on CIFAR-10 using ResNet56. The symbols (+) and (-) denote increase and decrease in accuracy regarding the original (unpruned) network, respectively. For each level of FLOP reduction ($\%$), we highlight the best results in bold and underline the second-best results.
	}
	\label{tab:main_resultsResNet56_left}
		\centering
		\scriptsize
		\renewcommand{\arraystretch}{1.3}
		\begin{tabular}{@{}l@{\hskip 13mm}c@{\hskip 13mm}c@{}}
			\hline
			Method                        & $\Delta$ Acc. & FLOPs\\ \hline
			DECORE~\cite{Alwani:2022} (CVPR, 2022)    & + 0.08         & 26.30      \\
			HALP~\cite{Shen:2022} (NeurIPS, 2022)	 & + 0.03		& 33.72 		\\
			SOKS~\cite{Liu:2023} (TNNLS, 2023)       & + 0.16         & \underline{35.91}      \\
			CKA~\cite{Pons:2024} (ICPR, 2024)          & \textbf{+ 1.25}         & \textbf{37.52}      \\ 
			Consensus (Ours)       & \underline{+ 0.99}         & \textbf{37.52}      \\ \hline
			GKP-TMI~\cite{Zhong:2022} (ICLR, 2022)     & + 0.22         & 43.23      \\
			GCNP~\cite{Jiang:2022} (IJCAI, 2022)      & + 0.13         & \underline{48.31}      \\
			CKA~\cite{Pons:2024} (ICPR, 2024)           & \textbf{+ 0.86}         & \textbf{48.78 }     \\
			Consensus (Ours)       & \underline{+ 0.72}         & \textbf{48.78}      \\ \hline
			RLAL~\cite{Ganjdanesh:2024}  (CVPR, 2024)    & + 0.45         & 50.00      \\
			UDSP~\cite{Gao:2024} (CVPR, 2024) & +0.66             & 50.10  \\
			GNN-RL~\cite{Yu:2022}  (ICML, 2022)    & + 0.10         & 54.00      \\
			ATO~\cite{Wu:2024} (CVPR, 2024)     & + 0.24         & 55.00      \\
			RL-MCTS~\cite{Wang:2022} (WACV, 2022)     & + 0.36         & 55.00      \\
			WhiteBox~\cite{WhiteBox:2023} (TNNLS, 2023) & + 0.28         & 55.60      \\
			CLR-RNF~\cite{Lin:2023} (TNNLS, 2023)    & + 0.01        & \underline{57.30}      \\
			CKA~\cite{Pons:2024} (ICPR, 2024)           & \textbf{+ 0.78}        & \textbf{60.04}      \\
			Consensus (Ours)       & \underline{+ 0.60}         & \textbf{60.04}      \\ \hline
			DAIS~\cite{Guan:2023} (TNNLS, 2023)      & - 1.00         & 70.90      \\
			HRank~\cite{Lin:2020} (CVPR, 2020)      & - 2.54         & 74.09      \\
			CKA~\cite{Pons:2024}  (ICPR, 2024)      & \textbf{+ 0.08}         & 75.05      \\
			Consensus$_{20}$ (Ours)       & \underline{- 0.17}         & 75.05      \\ 
			GCNP~\cite{Jiang:2022} (IJCAI, 2022)    & - 0.97         & \underline{77.22}      \\
			CKA~\cite{Pons:2024} (ICPR, 2024)           & - 0.66         & \textbf{78.80}      \\
			Consensus (Ours)       & - 0.58         & \textbf{78.80}      \\ \hline
		\end{tabular}
\end{table}

\begin{table}[H]
	\caption{Comparison with state-of-the-art pruning methods on ImageNet using ResNet50. The symbols (+) and (-) denote increase and decrease in accuracy regarding the original (unpruned) network, respectively. For each level of FLOP reduction ($\%$), we highlight the best results in bold and underline the second-best results.
	}
	\label{tab:main_resultsResNet56_rigth}
		\centering
		\scriptsize
		\renewcommand{\arraystretch}{1.3}
		\begin{tabular}{@{}l@{\hskip 13mm}c@{\hskip 13mm}c@{}}
			\hline
			Method                                      & $\Delta$ Acc.     & FLOPs          \\ \hline
			DECORE~\cite{Alwani:2022} (CVPR, 2022)      & + 0.16             & 13.45          \\
			SOSP~\cite{Nonnemacher:2022} (ICLR, 2022)   & + 0.41             & 21.00          \\
			GKP-TMI~\cite{Zhong:2022} (ICLR, 2022)      & - 0.19             & 22.50          \\
			CKA~\cite{Pons:2024} (ICPR, 2024)                                  & + 1.11            & 22.64          \\
			Consensus (Ours)       & \textbf{+ 2.04}         & 22.64      \\
			SOSP~\cite{Nonnemacher:2022} (ICLR, 2022)   & + 0.45             & \underline{28.00}          \\ 
			CKA~\cite{Pons:2024} (ICPR, 2024)                                  & + 0.74             & \textbf{28.30}          \\
			Consensus (Ours)       & \underline{+1.5}         & \textbf{28.30} \\ \hline
			GKP-TMI~\cite{Zhong:2022} (ICLR, 2022)      & - 0.62             & 33.74          \\
			LPSR~\cite{Zhang_Liu:2022} (SPL, 2022)    & - 0.57          & \underline{37.38}          \\
			CKA~\cite{Pons:2024} (ICPR, 2024)                                  & \textbf{- 0.18} & \textbf{39.62} \\
			Consensus (Ours)       & \underline{-0.35}         & \textbf{39.62} \\ \hline
			CLR-RNF~\cite{Lin:2023} (TNNLS, 2023)       & - 1.16             & 40.39          \\
			DECORE~\cite{Alwani:2022} (CVPR, 2022)      & - 1.57             & 42.30          \\
			HRank~\cite{Lin:2020} (CVPR, 2020)          & - 1.17             & 43.77          \\
			SOSP~\cite{Nonnemacher:2022} (ICLR, 2022)   & - 0.94             & 45.00          \\
			CKA~\cite{Pons:2024} (ICPR, 2024)  & - 0.90  & \underline{45.28}          \\
			Consensus (Ours)       & \underline{- 0.84}         & \underline{45.28} \\
			WhiteBox~\cite{WhiteBox:2023} (TNNLS, 2023) & \textbf{- 0.83}             & \textbf{45.60}          \\ \hline
			&          &       \\ 
			&          &       \\ 
			&          &       \\ 
			&          &       \\ 
			&          &       \\ 
			&          &       \\ 
		\end{tabular}
\end{table}
\vspace{-21mm}

While our method obtained comparable performance with CKA on CIFAR-10, on the more challenging ImageNet dataset, we outperformed it by a good margin. We believe the reason for these results is that, although CIFAR-10 is a popular dataset for benchmarking pruning methods, it is not as challenging as ImageNet. Table~\ref{tab:main_resultsResNet56_rigth} confirms this, where we outperform it by up to $0.93$ pp. Compared to other pruning techniques, our method exhibits a behavior similar to that on CIFAR-10: we obtain one of the best compromises between accuracy drop and FLOPs reduction.

\noindent
\textbf{Effectiveness of the Proposed Consensus Criterion.}
To confirm our intuition that a Consensus criterion is more effective than single similarity criteria, we conduct the following experiment.
%
First, we perform a single pruning iteration on ResNet32 using the individual metrics that compose our Consensus.
Then, for each criterion, we evaluate the accuracy of the obtained models on different adversarial attacks (CIFAR-10.2, CIFAR-C and FGSM) and clean images.

Following previous works~\cite{Jordao:2021}, for each criterion, we report the mean accuracy across all benchmarks. According to the results, we observe that our Consensus technique achieves the highest performance, outperforming the individual metrics by up to 0.60 pp. We highlight that, in some scenarios, individual metrics may surpass the performance of the Consensus criterion, but they are not consistent across all benchmarks compared to our method. 
These results reinforce that combining the strengths of multiple similarity metrics into a single criterion to guide the pruning process is a robust and reliable technique, offering many advantages, including mitigating shortcut opportunities that hinder generalization in more challenging scenarios such as adversarial attacks and out-of-distribution (OOD) samples~\cite{Geirhos:2020,Hermann:2023}.

The reason for the previous results is that our criterion carefully selects which structures, particularly layers, to eliminate from the architecture. To confirm this statement, we compare our criterion with existing layer pruning methods ranging from evolutionary algorithms~\cite{Zhou:2022}, projection methods~\cite{Jordao:2020}, Taylor expansion~\cite{Zhang_Liu:2022}, meta-learning~\cite{Chen:2019} and single similarity representation metrics~\cite{Pons:2024}.

Tables~\ref{tab:comparison_layer_methods_left} and \ref{tab:comparison_layer_methods_rigth} shows the results. The Consensus technique is either on par with or surpasses the performance of state-of-the-art layer pruning methods. Particularly at higher reduction levels, our method is able to maintain the predictive performance with low accuracy drop and, in some cases, even improve it compared to the unpruned model.

\begin{table}[!t]
	\caption{Comparison with state-of-the-art layer-pruning methods on CIFAR-10 using ResNet56. The symbols (+) and (-) denote increase and decrease in accuracy regarding the original (unpruned) network, respectively. For each level of FLOP reduction ($\%$), we highlight the best results in bold and underline the second-best results.
	}
	\label{tab:comparison_layer_methods_left}
	\centering
	\scriptsize
	\renewcommand{\arraystretch}{1.3}
		\begin{tabular}{@{}l@{\hskip 16mm}c@{\hskip 16mm}c@{}}
			\hline
			Method                        & $\Delta$ Acc. & FLOPs\\ \hline
			PLS~\cite{Jordao:2020} (J-STSP, 2020)    & - 0.98        & 30.00      \\
			FRPP~\cite{Chen:2019} (TPAMI, 2019)      & + 0.26      & 34.80      \\
			ESNB~\cite{Zhou:2022} (TNNLS, 2022)      & - 0.62     & 52.60      \\
			LPSR~\cite{Zhang_Liu:2022} (SPL, 2022) & + 0.19      & \underline{52.75}      \\
			CKA~\cite{Pons:2024} (ICPR, 2024)             & \textbf{+ 0.95}         & \textbf{56.29}      \\
			Consensus (ours)    & \underline{+ 0.50}      & \textbf{56.29}      \\ \hline
			PLS~\cite{Jordao:2020} (J-STSP, 2020)     & - 0.91     & 62.69      \\
			LPSR~\cite{Zhang_Liu:2022} (SPL, 2022) & - 0.87     & 71.65      \\
			CKA~\cite{Pons:2024} (ICPR, 2024)             & \textbf{+ 0.16}         & \underline{71.30}      \\
			Consensus (ours)    & \underline{+ 0.09}      & \underline{71.30}      \\ \hline
			CKA~\cite{Pons:2024} (ICPR, 2024)             & + 0.08         & \textbf{75.05}      \\
			Consensus (ours)    & - 0.17      & \textbf{75.05}      \\ \hline
		\end{tabular}
\end{table}
		\begin{table}[!t]
			
			\caption{Comparison with state-of-the-art layer-pruning methods on ImageNet using ResNet50. The symbols (+) and (-) denote increase and decrease in accuracy regarding the original (unpruned) network, respectively. For each level of FLOP reduction ($\%$), we highlight the best results in bold and underline the second-best results.
			}
			\label{tab:comparison_layer_methods_rigth}
			\centering
			\scriptsize
			\renewcommand{\arraystretch}{1.3}
			\begin{tabular}{@{}l@{\hskip 17mm}c@{\hskip 17mm}c@{}}
			\hline
			Method                                      & $\Delta$ Acc.     & FLOPs          \\ \hline
			CKA~\cite{Pons:2024} (ICPR, 2024)                                  & \underline{+ 1.11}            & \textbf{22.64}          \\
			Consensus (Ours)       & \textbf{+ 2.04}         & \textbf{22.64}      \\  \hline
			LPSR~\cite{Zhang_Liu:2022} (SPL, 2022)      & - 1.38             & \underline{37.38}          \\
			CKA~\cite{Pons:2024} (ICPR, 2024)                 & \textbf{- 0.18}          & \textbf{39.62}          \\
			Consensus (Ours)          & \underline{- 0.35} & \textbf{39.62} \\ \hline
			PLS~\cite{Jordao:2020} (J-STSP, 2020) & \textbf{- 0.67} & \textbf{45.28} \\
			CKA~\cite{Pons:2024} (ICPR, 2024)                 & - 0.90         & \textbf{45.28}          \\
			Consensus (Ours)          & \underline{- 0.84}          & \textbf{45.28} \\ \hline
			&          &       \\ 
			&          &       \\ 
			&          &       \\ 
			&          &       \\ 
			
		\end{tabular}
	\vspace{-15mm}
\end{table}
\noindent
\textbf{Adversarial Robustness of the Proposed Consensus Criterion.}
To demonstrate the efficacy of our proposed Consensus criterion in adversarial scenarios, we conduct a comprehensive evaluation focusing on the robustness of the pruned models. To this end, we utilize four widely recognized benchmarks: CIFAR-10.2, CIFAR-C, ImageNet-C, and the Fast Gradient Sign Method (FGSM) attack. These benchmarks encompass a variety of adversarial attacks and perturbation scenarios, providing a thorough assessment of the pruned models robustness.

\begin{figure}[!ht]
	\centering
	\includegraphics[width=0.49\linewidth]{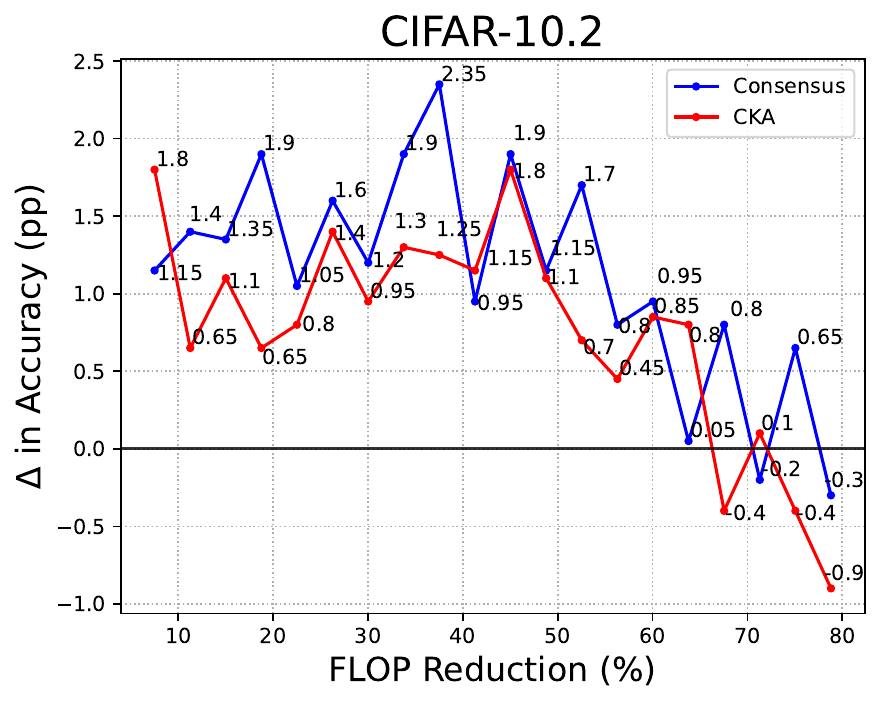}
	\includegraphics[width=0.48\linewidth]{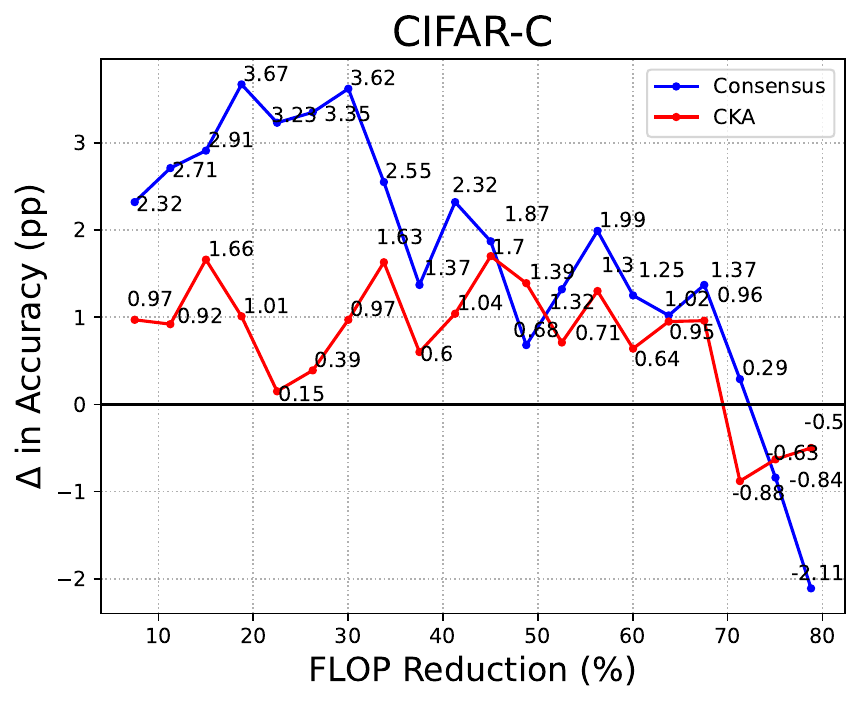}
	\includegraphics[width=0.477\linewidth]{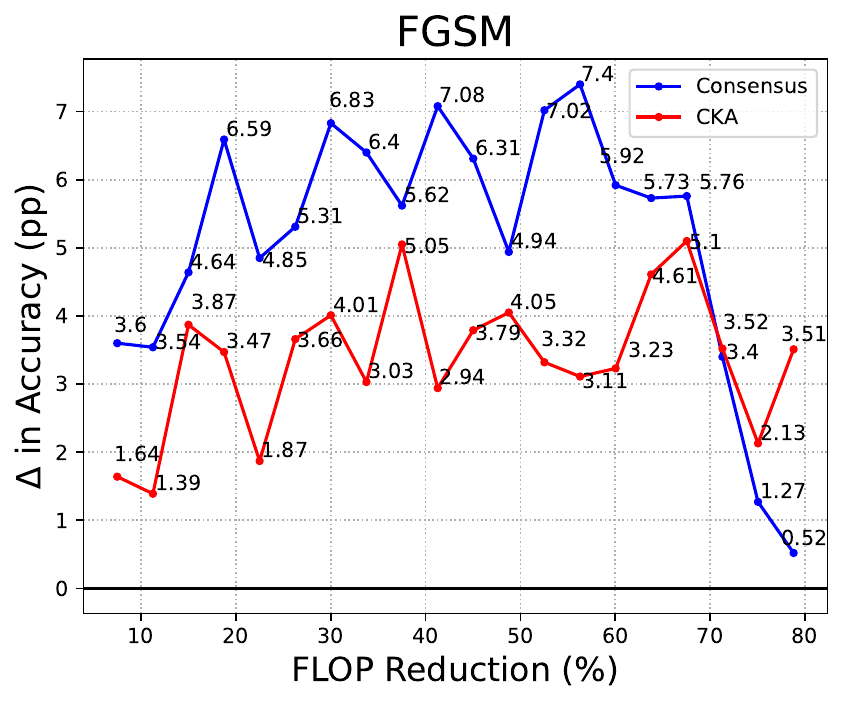}
	\includegraphics[width=0.49\linewidth]{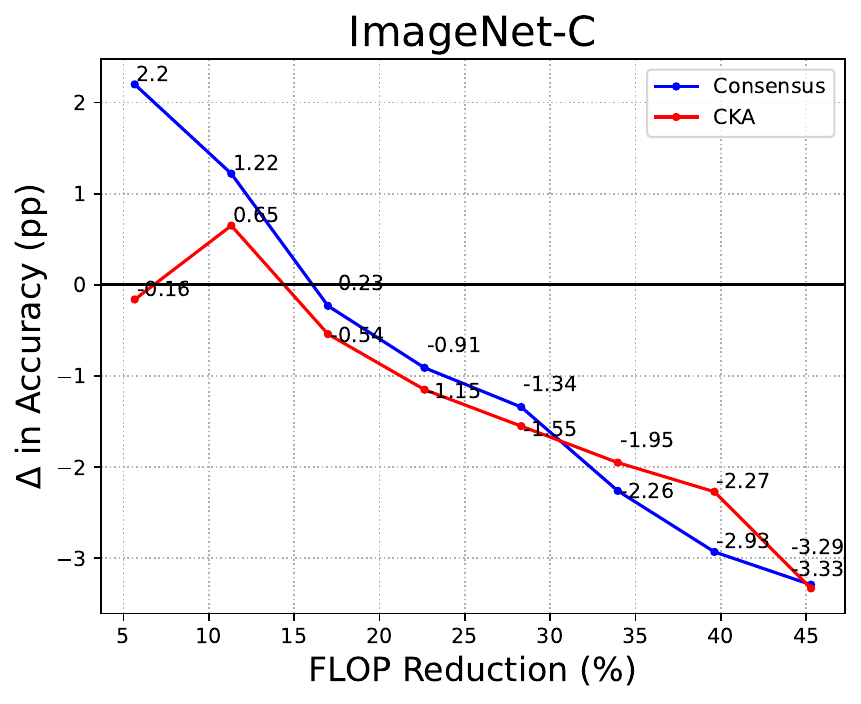}
	\caption{Trade-offs between predictive accuracy and computational performance (FLOP reduction) of pruned models under different types of adversarial attacks and out-of-distribution (OOD) samples. The figures show the change in accuracy (indicated as $\Delta$ in Accuracy) compared to the original, unpruned, model as a function of FLOP reduction for four different benchmarks: CIFAR-10.2 (OOD samples, top-left), CIFAR-C (semantic-preserving attack, top-right), FGSM (adversarial perturbation attack, bottom-left) and ImageNet-C (semantic-preserving attack, botton-right). Blue curves indicate the accuracy at different compression rates using our Consensus method. The red curves represent the accuracy using CKA by Pons et al.~\cite{Pons:2024}, a promising layer-pruning technique.}
	\label{fig:adversarial_attacks}
\end{figure}

Figure~\ref{fig:adversarial_attacks} (top-left) shows the results of the Out-of-Distribution scenario, CIFAR-10.2. Our method outperformed CKA in most cases including on the highest compression rate (i.e., above $70\%$). In particular, at the same compression rate, we obtain an improvement of up to $1.20$ pp.

On the FGSM attack (Figure~\ref{fig:adversarial_attacks}, bottom-left), our method surpasses CKA by a large margin. More concretely, at a FLOP reduction of $41.28\%$ and $56.29\%$, CKA underperformed ours by 4.14 and 4.29 pp. 
We observe a similar behavior when comparing the methods on the semantic-preserving attacks, CIFAR-C (Figure~\ref{fig:adversarial_attacks}, top-right) and ImageNet-C (Figure~\ref{fig:adversarial_attacks}, bottom-right). We reinforce that on ImageNet without adversarial attacks (Table~\ref{tab:main_resultsResNet56_rigth}), our criterion also provides better results in terms of accuracy drop.

Overall, the previous discussion confirms that by integrating multiple similarity criteria, our method effectively mitigates the limitations of single-metric approaches, leading to enhanced robustness against adversarial examples.

Even though our method outperforms CKA, it is important to observe that CKA promotes pruned models less sensitive to attacks compared to the original, unpruned, model. This evidence suggests that the similarity metric is a promising line of research in the context of pruning.

We also evaluate the adversarial robustness of the Consensus method against the pruned models defined by Jordao et al.~\cite{Jordao:2021}. In their work, the authors investigated the effectiveness of pruned models as adversarial defense mechanisms.
The goal is to verify changes in the adversarial robustness of pruned models after a single pruning iteration. Compared to the results by Jordao et al.~\cite{Jordao:2021}, our Consensus technique exhibits an improvement of 1.48 pp and 0.84 pp on the FGSM and CIFAR-C benchmarks, respectively. On challenging ImageNet-C, Consensus demonstrates an improvement of 1 pp. 
Although filter pruning is beyond the scope of this work. When we compare the pruned models obtained through a filter pruning process, our method achieves an accuracy improvement of up to 1.18 pp, showing gains on all available benchmarks.
These results confirm that our Consensus technique delivers remarkable performance, outperforming filter pruning methods while also inheriting all the benefits of layer pruning.

\noindent
\textbf{Effectiveness in Shallow Architectures.}
Although modern models rely on the \emph{we need to go deeper} paradigm, shallow models still play an important role in downstream tasks~\cite{Touvron:2023}.
Particularly, shallow models are more attractive in low-resource scenarios, and applying pruning to them leads to even better performance. On the other hand, due to their low capacity, shallow models may be more sensitive to pruning.
In this experiment, we assess the effectiveness of our method in pruning shallow models. For this purpose, we compare our method with state-of-the-art pruning techniques on the ResNet32 and ResNet44 architectures. Table~\ref{tab:shallow_models} summarizes the results.

\begin{table}[!hbt]
	\centering
	\scriptsize
	\renewcommand{\arraystretch}{1.3}
	\caption{Comparison of state-of-the-art pruning methods on CIFAR-10 using ResNet32 and ResNet44. The symbols (+) and (-) denote increase and decrease in accuracy regarding the original (unpruned) network, respectively. For each level of FLOP reduction ($\%$), we highlight the best results in bold and underline the second best results.}
	\label{tab:shallow_models}
	\begin{tabular}{llrc}
		\hline
		& Method                             & $\Delta$ Acc. & FLOPs (\%)     \\ \hline
		\multicolumn{1}{l|}{\multirow{11}{*}{ResNet32}} & GKP-TMI~\cite{Zhong:2022} (ICLR, 2022)    & + 0.22           & 43.10 \\
		\multicolumn{1}{l|}{} & SOKS~\cite{Liu:2023} (TNNLS, 2023) & - 0.38          & \underline{46.85}          \\
		\multicolumn{1}{l|}{} & CKA~\cite{Pons:2024} (ICPR, 2024)                         & \textbf{+ 0.68} & \textbf{47.78} \\
		\multicolumn{1}{l|}{} & Consensus (ours)                         & \underline{+ 0.56} & \textbf{47.78} \\ \cline{2-4} 
		\multicolumn{1}{l|}{} & DAIS~\cite{Guan:2023} (TNNLS, 2023)       & \textbf{+ 0.57} & 53.90 \\
		\multicolumn{1}{l|}{} & SOKS~\cite{Liu:2023} (TNNLS, 2023) & - 0.80          & \underline{54.58}          \\
		\multicolumn{1}{l|}{} & CKA~\cite{Pons:2024} (ICPR, 2024)                         & + 0.05          & \textbf{54.61} \\
		\multicolumn{1}{l|}{} & Consensus (ours)                         & \underline{+ 0.16}          & \textbf{54.61} \\ \cline{2-4} 
		\multicolumn{1}{l|}{} & CKA~\cite{Pons:2024} (ICPR, 2024)                         & \underline{- 0.18} & \underline{61.44} \\ 
		\multicolumn{1}{l|}{} & Consensus (ours)                         & \textbf{- 0.11} & \underline{61.44} \\
		\multicolumn{1}{l|}{} & SOSP~\cite{Nonnemacher:2022} (ICLR, 2022) & - 0.24          & \textbf{67.36} \\ \hline
		\multicolumn{1}{l|}{\multirow{8}{*}{ResNet44}} & AGMC~\cite{Yu:2021} (ICCV, 2021)                  & - 0.82          & 50.00          \\
		\multicolumn{1}{l|}{}  & DCP-CAC~\cite{Chen:2021} (TNNLS, 2022)                     & - 0.03          & \underline{50.04} \\
		\multicolumn{1}{l|}{} & CKA~\cite{Pons:2024} (ICPR, 2024)                         & \underline{+ 0.47} & \textbf{53.27} \\
		\multicolumn{1}{l|}{} & Consensus (ours)                         & \textbf{+ 0.63} & \textbf{53.27} \\ \cline{2-4} 
		\multicolumn{1}{l|}{} & CKA~\cite{Pons:2024} (ICPR, 2024)                         & \underline{+ 0.22}          & \textbf{62.95}          \\
		\multicolumn{1}{l|}{} & Consensus (ours)                         & \textbf{+ 0.50}          & \textbf{62.95}          \\ \cline{2-4}
		\multicolumn{1}{l|}{} & CKA~\cite{Pons:2024} (ICPR, 2024)                         & \underline{- 0.29}          & \textbf{72.64}          \\ 
		\multicolumn{1}{l|}{} & Consensus (ours)                         & \textbf{- 0.08}          & \textbf{72.64}          \\ \hline
		
	\end{tabular}
	\vspace{-4mm}
\end{table}

From Table~\ref{tab:shallow_models}, we highlight the following key observations: On both architectures, our method notably outperforms CKA as we increase the FLOP reduction, except for ResNet32 at the lower compression rate. For ResNet32, the Consensus method achieves an accuracy improvement of 0.56 with a 47.78\% reduction in FLOPs. For ResNet44, it achieves an improvement of 0.63 with a 53.27\% reduction in FLOPs. Such a finding reinforces that, besides achieving superior robustness (see Figure~\ref{fig:adversarial_attacks}), a consensus of criteria favors identifying unimportant layers in shallow architectures. Moreover, our results are on par with or superior to top-performing approaches for shallow architectures.

\noindent
\textbf{GreenAI and Computational Costs.} 
The concept of GreenAI has gained significant attention in the research community, emphasizing the need for more environmentally friendly AI practices by reducing the computational resources required for training and deploying models~\cite{Faiz:2024, Schwartz:2020}. Our layer-pruning technique aligns with this vision by significantly improving the computational costs associated with large models. 
Specifically, our pruned models achieve substantial reductions in latency and FLOPs, thereby decreasing the energy consumption during model training and inference.
These improvements directly translate into lower carbon emissions, as our (pruned) models require fewer computational resources for training and fine-tuning processes~\cite{Strubell:2019}.
As a concrete example, on ResNet56, we achieve a reduction of approximately 68.75\% in carbon emissions and 66.99\% in financial costs\footnote{For reproducibility purposes, we estimate these values using the MachineLearning Impact calculator~\cite{Lacoste:2019} and the vast.ai GPU usage prices.}. These findings reinforce the potential of advanced pruning techniques in promoting sustainable AI practices, minimizing environmental impact, and contributing to the GreenAI initiative~\cite{Strubell:2019, Lacoste:2019}.

\noindent
\textbf{Effectiveness in Transformer Architectures.}
Recent progress in foundation models frequently relies on Transformer architectures and their variations~\cite{Touvron:2023}. Our study investigates the effectiveness of the Consensus method on the widely adopted Transformer architecture. Due to limited computational resources, we follow Pons et al.~\cite{Pons:2024} and limit our analysis to tabular data, as Visual Transformers typically require larger datasets to achieve results on pair with convolutional networks.  It is important to mention that our goal here is not to advance the state-of-the-art but rather to verify the effectiveness of our layer-pruning technique in Transformer architectures.

Our Transformer consists of 10 layers, each with 128 heads and projection dimension of 64. As in previous experiments, we conduct a fine-tuning process after each pruning iteration. We evaluate the effectiveness of our layer-pruning technique on Transformers for human activity recognition based on wearable sensors, a popular application involving tabular data. Details about these datasets are available in the work by Sena et al.~\cite{SENA:2021}.

\begin{figure}[H]
	\centering
	\includegraphics[width=0.49\linewidth]{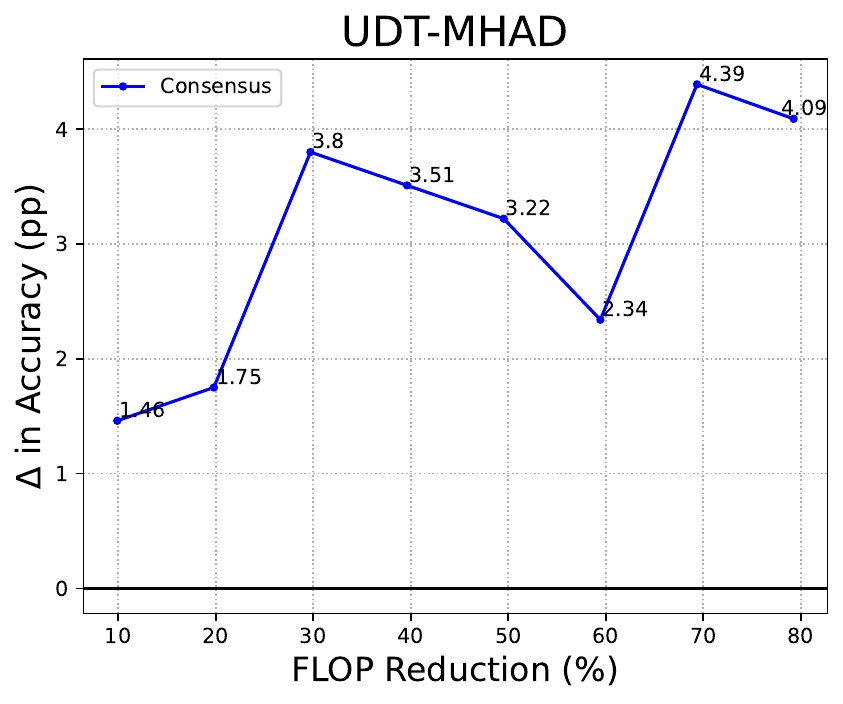}
	\includegraphics[width=0.49\linewidth]{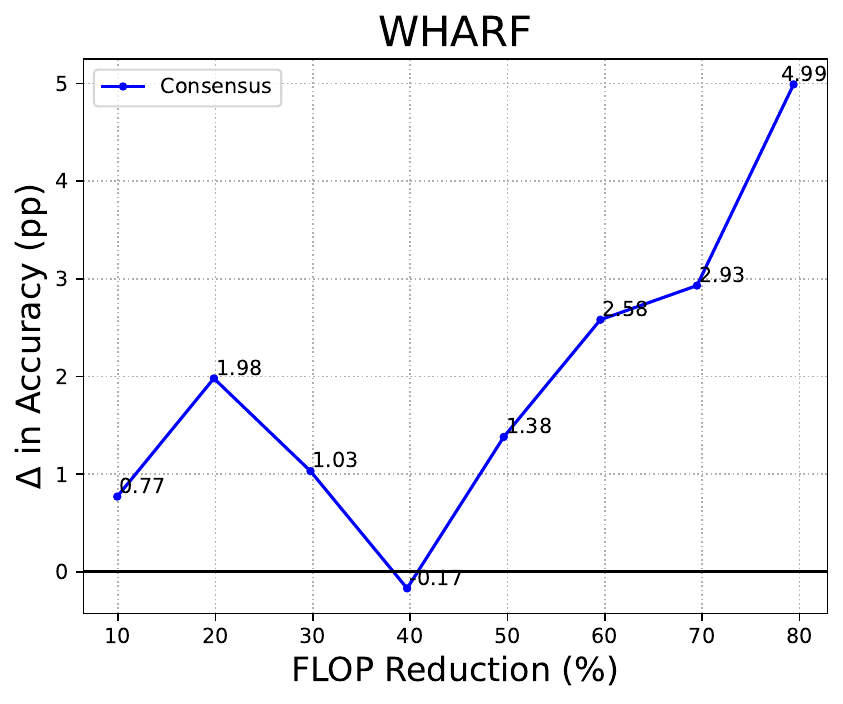}
	\includegraphics[width=0.49\linewidth]{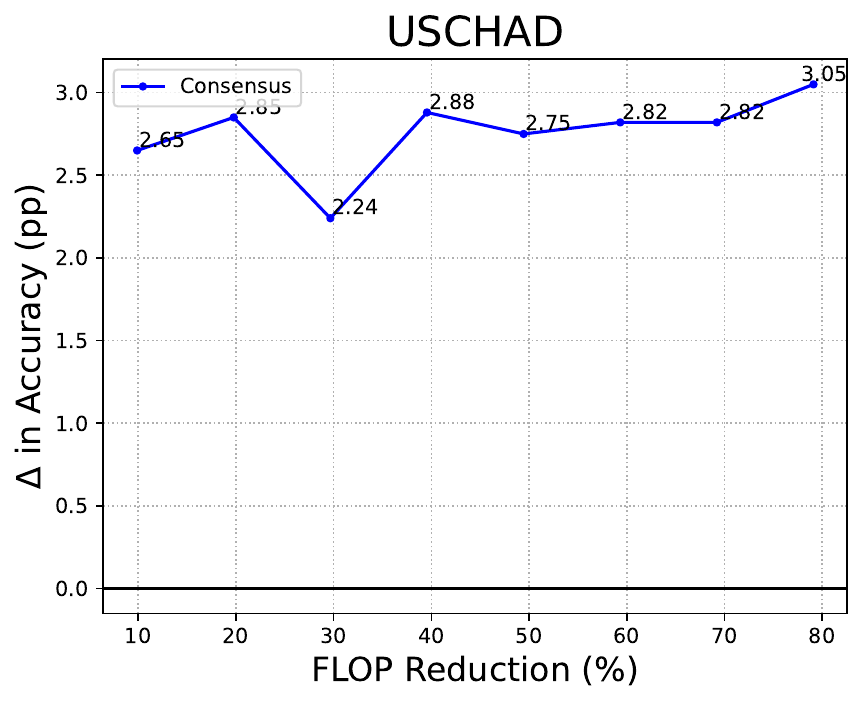}
	\includegraphics[width=0.49\linewidth]{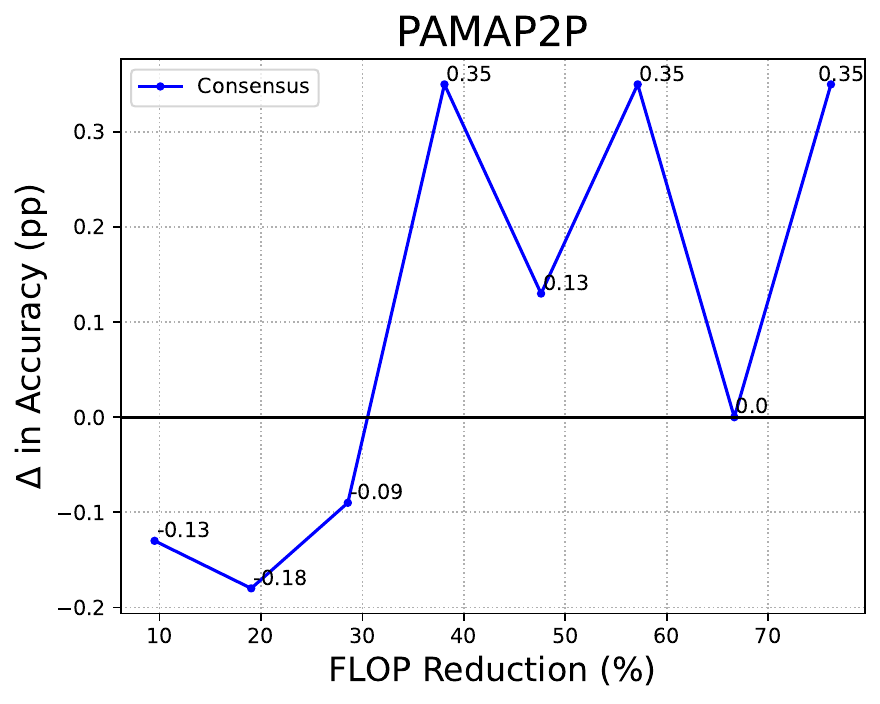}
	\caption{Performance of the Consensus criterion on Transformer architecture for human activity recognition based on wearable sensors (tabular data). The blue curves denote the performance of the pruned models. Positive values indicate an improved accuracy compared to the original, unpruned, model.}
	\label{fig:transformers_performance}
\end{figure}

Figure~\ref{fig:transformers_performance} shows the results. The solid black line indicates the point where the accuracy drop is zero; thus, pruned models above or below this line exhibit an improvement or deterioration in accuracy, respectively. We observe that our pruning technique reduces FLOPs by up to $80\%$ with a negligible drop in accuracy, thus confirming its effectiveness in the Transformers architectures. 
Furthermore, as the Consensus technique extracts features using $M(\cdot, X)$, pruning effectiveness varies across datasets, showing dependence on data nature and Transformer model architecture. 

In summary, our Consensus criterion enhances the efficiency of Transformer models in human activity recognition based on tabular data, with effects dependent on the degree of pruning and the specific dataset.


%
\section{Conclusions}\label{sec:conclusions}
Layer pruning is a technique that excels in model compression and acceleration. However, existing criteria for layer selection may not fully capture the underlying properties of these structures. Our approach advances the field by more comprehensively addressing the limitations of existing pruning techniques that employ single metrics. These metrics often struggle to effectively capture the underlying properties of layers and are prone to the phenomenon of shortcut learning, where models tend to exploit undesirable shortcuts in training data that poorly generalize to new situations. Our Consensus technique preserves model accuracy even at high compression rates, significantly improves computational performance by reducing inference time and memory consumption, and increases robustness to adversarial attacks, providing a more robust evaluation of layer importance. 
Extensive experiments on standard benchmarks and architectures confirm the effectiveness of our method, achieving state-of-the-art performance in terms of preserving accuracy, FLOPs reduction and adversarial robustness, thereby achieving a triple-win outcome. 
Specifically, we reduce FLOPs by up to 78.8\% with minimal accuracy loss and improve adversarial robustness by up to 4 percentage points compared to state-of-the-art methods. Additionally, our results highlight the benefits for GreenAI, with significant reductions 68.75\% in carbon emissions required for the training and fine-tuning of modern architectures.
\section*{Acknowledgments}

The authors would like to thank grant \#2023/11163-0, S\~ao Paulo Research Foundation (FAPESP), and grant \#402734/2023-8, National Council for Scientific and Technological Development (CNPq). Artur Jordao Lima Correia would like to thank Edital Programa de Apoio aos Novos Docentes 2023. Processo USP nº: 22.1.09345.01.2. Anna H. Reali Costa would like to thank grant \#312360/2023-1 CNPq. This study was also partially financed by the Coordenação de Aperfeiçoamento de Pessoal de Nivel Superior – Brasil (CAPES) – Finance Code 001.


{
\bibliographystyle{ieee_fullname}
\bibliography{refs}
}

\clearpage

\end{document}